\documentclass{article}

\usepackage[final,main]{neurips_2026}

\usepackage[utf8]{inputenc}
\usepackage[T1]{fontenc}
\usepackage{hyperref}
\usepackage{url}
\usepackage{booktabs}
\usepackage{amsfonts}
\usepackage{amsmath}
\usepackage{amssymb}
\usepackage{nicefrac}
\usepackage{microtype}
\usepackage{xcolor}
\usepackage{graphicx}
\usepackage{tikz}
\usepackage{algorithm}
\usepackage{algorithmicx}
\usepackage{algpseudocode}

\usetikzlibrary{arrows.meta, shapes, positioning, decorations.pathreplacing, fit, calc}
\usepackage{pgfplots}
\pgfplotsset{compat=1.18}
\usepackage{amsthm}
\newtheorem{remark}{Remark}

\title{Scalable Hyperparameter-Divergent Ensemble Training with Automatic Learning Rate Exploration for Large Models}

\author{%
  Hailing Cheng \\
  LinkedIn Inc.\\
  Mountain View, CA 94043 \\
  \texttt{haicheng@linkedin.com} \\
  \And
  Tao Huang \\
  LinkedIn Inc.\\
  Mountain View, CA 94043 \\
  \texttt{taohuang@linkedin.com} \\
  \And
  Chen Zhu \\
  LinkedIn Inc.\\
  Mountain View, CA 94043 \\
  \texttt{chzhu@linkedin.com} \\
  \And
  Antonio Alonso \\
  LinkedIn Inc.\\
  Mountain View, CA 94043 \\
  \texttt{aalonso@linkedin.com} \\
}

\begin{document}

\maketitle

\begin{abstract}
Training large neural networks with data-parallel stochastic gradient descent
allocates $N$ GPU replicas to compute effectively identical updates---a practice that
leaves the rich space of learning rate configurations entirely unexplored during training.
We propose \textbf{Hyperparameter-Divergent Ensemble Training (HDET)}, a method that
repurposes these replicas for simultaneous learning rate exploration at negligible
communication overhead.
HDET operates in alternating phases: a \emph{fan-out} stage in which replicas train
independently under a structured, symmetric spread of learning rates, and a
\emph{converge} stage in which parameters are averaged across all replicas via
AllReduce every $T$ steps.
Building on this ensemble substrate, we further propose an
\textbf{automatic learning rate (auto-LR) controller} that treats the relative
training loss across replicas as a performance signal, updating the shared base
schedule toward higher-performing configurations via a momentum-based
gradient-free meta-update.
The combined method produces a self-adapting learning rate schedule that improves
both optimization quality and generalization without additional hyperparameter sweeps
or training budget.

Crucially, the framework generalizes beyond learning rate: any scalar hyperparameter
that does not alter model architecture---such as dropout rate, attention scale
temperature, or weight-decay coefficient---can be explored across replicas using the
same fan-out/converge protocol, with inter-replica loss differences serving as
zero-order \emph{hypergradients} that guide the search direction.
HDET is implemented as a drop-in replacement for PyTorch's \texttt{OneCycleLR}
scheduler, requiring no changes to model architecture, optimizer, or data pipeline.
Code is available at \url{https://github.com/hailingc/ensemble_training}.
\end{abstract}

\section{Introduction}
\label{sec:intro}

The learning rate schedule is among the most consequential hyperparameter choices
in large-model training.  Popular schedules---one-cycle annealing
\citep{smith2019super}, cosine decay \citep{loshchilov2017sgdr}, and linear
warmup-decay \citep{vaswani2017attention}---must be specified before training begins,
leaving practitioners to rely on expensive grid searches or heuristics calibrated
on smaller proxy models.  Even when a good schedule is found, it may not remain
optimal as model or dataset scale changes across training runs.

Simultaneously, large-model training routinely parallelizes across hundreds of GPU
replicas using data-parallel stochastic gradient descent (DP-SGD).  In the standard
paradigm, every replica maintains an identical copy of the model parameters and
advances under an identical learning rate; the replica's sole distinguishing role is
the mini-batch it processes.  Beyond improving gradient estimates through a larger
effective batch, the additional hardware contributes \emph{no qualitative diversity}
to the optimization trajectory.

We observe that this uniformity is a missed opportunity.  The $N$ parallel replicas
could explore $N$ distinct learning rate trajectories simultaneously, at no
additional hardware cost.  Moreover, if those diverging trajectories periodically
merge through parameter averaging, the resulting model can harvest
ensemble-like generalization benefits within a \emph{single} training run.

We formalize this intuition in \textbf{Hyperparameter-Divergent Ensemble Training
(HDET)}, which interleaves two phases atop any existing DDP+OneCycleLR setup:

\begin{enumerate}
\item \textbf{Fan-out}: each of $N$ GPU replicas trains under a distinct learning
  rate drawn from a symmetric spread $[\bar\eta\,(1-\alpha),\; \bar\eta\,(1+\alpha)]$
  around the current base schedule $\bar\eta^{(t)}$, where $\alpha \geq 0$ is the
  spread ratio;
\item \textbf{Converge}: every $T$ training steps, parameters are averaged across
  all replicas via AllReduce, producing a unified model that seeds the next
  fan-out round.
\end{enumerate}

On top of this ensemble substrate we introduce an \textbf{auto-LR controller}
that uses the relative training loss across replicas as a performance signal.
Replicas with below-average loss receive proportionally higher weight in a
softmax-weighted average of per-rank learning rates; the resulting performance
deviation drives a momentum-based velocity that shifts the base schedule up or down
at each sync point.  If high-LR replicas consistently outperform low-LR ones,
the base schedule adapts upward; if the opposite holds, it adapts downward.
This constitutes a gradient-free, within-run optimizer over the learning rate
schedule, eliminating the need for a priori schedule selection.

\paragraph{Contributions.}
\begin{itemize}
  \item We propose HDET, a training method that transforms existing DDP replicas
    into a structured learning rate exploration ensemble at zero hardware overhead
    (\S\ref{sec:method}).
  \item We introduce an automatic LR controller that uses rank-wise loss signals
    to adaptively update the learning rate via a gradient-free momentum update,
    requiring no additional tuning (\S\ref{sec:autolr}).
  \item We identify a practical two-stage strategy---warm noisy initialization from a
    pre-trained checkpoint followed by HDET ensemble training---that jointly maximizes
    model quality and training stability: warm initialization seeds each replica near a
    high-quality solution, while periodic weight averaging prevents divergence under
    aggressive learning rates that would destabilize either technique alone
    (\S\ref{sec:warminit}, \S\ref{sec:experiments}).
  \item We demonstrate empirically that HDET consistently improves final model
    quality and convergence speed on production-scale training tasks
    (\S\ref{sec:experiments}).
  \item We release a reference implementation as a drop-in PyTorch scheduler
    at \url{https://github.com/hailingc/ensemble_training}.
\end{itemize}

\section{Related Work}
\label{sec:related}

\paragraph{Learning-rate scheduling and online adaptation.}
Standard schedules---one-cycle \citep{smith2019super}, cyclical \citep{smith2017cyclical},
cosine annealing \citep{loshchilov2017sgdr}, and warmup-decay \citep{vaswani2017attention}---must
be fixed before training; \texttt{OneCycleLR} \citep{paszke2019pytorch} is the base
scheduler HDET subclasses.
Online methods eliminate this: Hypergradient Descent \citep{baydin2018hypergradient},
L4 \citep{rolinek2018l4}, D-Adaptation \citep{defazio2023learning}, DoG
\citep{ivgi2023dog}, and Schedule-Free \citep{defazio2024road} all derive step sizes
from gradient statistics or loss dynamics.
HDET differs: its adaptation signal is \emph{cross-replica performance} rather than
gradient information.

\paragraph{Learning-rate diversity and population-style exploration.}
Alrao \citep{blier2019learning} assigns different learning rates \emph{within} a
single network, covering a wide LR range at negligible cost.
PBT \citep{jaderberg2017population} maintains a population of models, periodically
evaluating and copying hyperparameters from better to worse members.
HDET shares concurrent hyperparameter exploration but reuses existing DDP replicas
and merges them by AllReduce averaging rather than replacement, requiring no extra
workers and preserving gradient information from every replica.

\paragraph{Distributed data-parallel training.}
Standard DP-SGD \citep{li2020pytorch} synchronizes gradients via AllReduce, keeping
all ranks at identical parameters $\theta_r^{(t)} = \theta^{(t)}$.
HDET retains gradient synchronization but assigns rank-specific LR multipliers,
letting parameters diverge between sync intervals.

\paragraph{Weight averaging and ensemble merging.}
SWA \citep{izmailov2018averaging}, EMA \citep{ho2020denoising}, and Model Soups
\citep{wortsman2022model} average model weights sequentially or after training to reach
flatter minima or improve generalization.
SWAP \citep{gupta2020stochastic} extends SWA to parallel workers trained simultaneously,
the setting most similar to HDET; however, SWAP workers share identical hyperparameters
and differ only in effective batch size, whereas HDET deliberately assigns distinct
learning rates across replicas to explore the hyperparameter space before each
averaging step.

\paragraph{Worker divergence and periodic synchronization.}
EASGD \citep{zhang2015easgd}, Local SGD \citep{haddadpour2019local}, DiLoCo
\citep{douillard2023diloco}, and GRAWA \citep{dimlioglu2024grawa} all allow workers
to drift from a shared solution between sync events to reduce communication or improve
convergence.
HDET differs: divergence is \emph{intentionally induced} by a structured LR spread
while standard synchronized-gradient DDP is retained throughout.

\paragraph{Gradient-free population optimization.}
NES \citep{wierstra2014natural} and ES \citep{salimans2017evolution} optimize model
parameters via a population of candidates and scalar fitness signals, requiring no
gradient information.
HDET's auto-LR controller borrows this population-signal idea narrowly: inter-replica
losses steer a single shared scalar (the base LR) while standard gradient-based weight
updates remain intact.

\paragraph{Deep ensembles.}
Deep Ensembles \citep{lakshminarayanan2017simple} capture epistemic uncertainty by
training $N$ independent models from different initializations, at $N\times$ cost;
diversity arises primarily from models reaching different loss basins
\citep{fort2019deep}.
HDET generates comparable diversity within a single run by using the LR spread to
push replicas toward different basins before averaging.

\section{Hyperparameter-Divergent Ensemble Training}
\label{sec:method}

\subsection{Overview}

\begin{figure}[t]
\centering
\begin{tikzpicture}[
  >=Latex,
  syncnode/.style={draw, rounded corners=4pt, fill=orange!20,
                   minimum width=1.8cm, minimum height=0.65cm,
                   font=\small, align=center},
  gpubox/.style={draw, rounded corners=2pt, fill=blue!10,
                 minimum width=2.2cm, minimum height=0.65cm,
                 font=\scriptsize, align=center},
  lbl/.style={font=\scriptsize\itshape, color=gray!60!black}
]

\node[syncnode] (s0) at (0,0) {Sync\\$\bar\theta^{(kT)}$};

\node[gpubox] (g0) at (4.2, 2.0) {GPU 0 \quad $\eta_0 = (1{-}\alpha)\bar\eta$};
\node[gpubox] (g1) at (4.2, 0.8) {GPU 1 \quad $\eta_1$};
\node[gpubox] (g2) at (4.2,-0.4) {GPU 2 \quad $\eta_2$};
\node[gpubox] (g3) at (4.2,-1.6) {GPU 3 \quad $\eta_3 = (1{+}\alpha)\bar\eta$};

\draw[->] (s0.east) -- (g0.west);
\draw[->] (s0.east) -- (g1.west);
\draw[->] (s0.east) -- (g2.west);
\draw[->] (s0.east) -- (g3.west);

\draw[decorate,decoration={brace,amplitude=4pt,mirror}]
  (3.0,-2.0) -- (5.5,-2.0)
  node[midway, below=5pt, lbl] {$T$ gradient steps (shared gradient, divergent $\eta_r$)};

\node[syncnode] (s1) at (8.5, 0) {Sync\\$\bar\theta^{((k+1)T)}$};

\draw[->] (g0.east) -- (s1.west);
\draw[->] (g1.east) -- (s1.west);
\draw[->] (g2.east) -- (s1.west);
\draw[->] (g3.east) -- (s1.west);

\node[above=0.25cm of s1, lbl, align=center]
  {AllReduce (mean):\\$\bar\theta' = \frac{1}{N}\sum_r \theta_r$};

\draw[->, dashed] (s1.east) -- ++(0.8,0)
  node[right, lbl, align=left] {reshuffle $\eta_r$\\next cycle};

\end{tikzpicture}
\caption{%
  One HDET fan-out/converge cycle with $N{=}4$ replicas and spread ratio $\alpha$.
  All replicas share the same all-reduced gradient at each step but take steps of
  different sizes, causing parameters to diverge.  Every $T$ steps an AllReduce
  average collapses the ensemble into a single model.  Learning rates are randomly
  reassigned (reshuffled) before the next cycle.
}
\label{fig:cycle}
\end{figure}
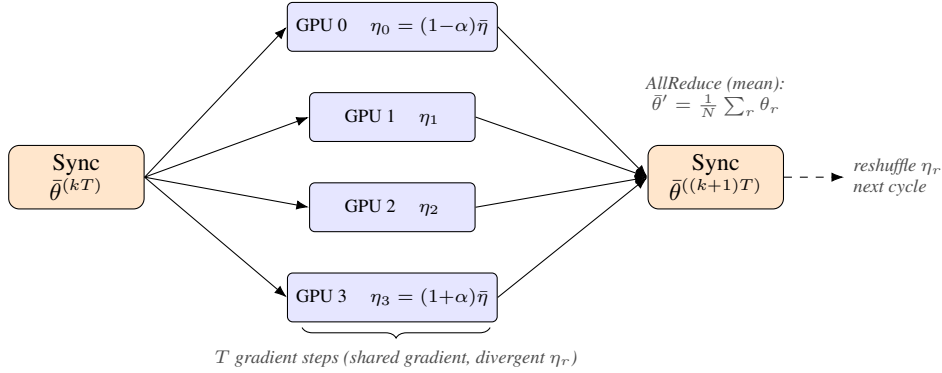

HDET adds three targeted modifications to standard DDP+OneCycleLR training:
(1) a structured LR spread across replicas at initialization;
(2) periodic all-reduce weight averaging every $T$ steps;
and (3) an optional auto-LR controller that adjusts the base schedule between sync
intervals.  Figure~\ref{fig:cycle} illustrates the full fan-out/converge cycle.

\subsection{Learning Rate Divergence (Fan-Out)}
\label{sec:fanout}

For $N$ ranks and spread ratio $\alpha \geq 0$, define
$\delta = \max\!\left(\tfrac{N-1}{2},\, 0.5\right)$ and assign each rank $r$ the multiplier
\begin{equation}
  \rho_r = 1 + \alpha \cdot \frac{r - \frac{N-1}{2}}{\delta},
  \label{eq:rho}
\end{equation}
so that $\sum_r \rho_r = N$ (symmetric around 1) and $\rho_r \in [1{-}\alpha,\,1{+}\alpha]$.
Rank $r$ trains under $\eta_r^{(t)} = \eta^{(t)} \cdot \rho_r$, where $\eta^{(t)}$
is the shared OneCycleLR base schedule, but all ranks share the same AllReduce gradient:
\begin{equation}
  \theta_r^{(t+1)} = \theta_r^{(t)} - \eta_r^{(t)} \cdot \bar{g}^{(t)}.
  \label{eq:rank_lr}
\end{equation}
Setting $\alpha{=}0$ recovers standard DDP exactly.

\subsection{Weight Averaging (Converge)}
\label{sec:converge}

Every $T$ steps, HDET AllReduce-averages all model parameters across ranks:
\begin{equation}
  \theta_r^{(t)} \;\leftarrow\; \frac{1}{N}\sum_{r'=0}^{N-1} \theta_{r'}^{(t)},
  \qquad \forall\, r.
  \label{eq:avg}
\end{equation}
By SWA theory \citep{izmailov2018averaging}, averaged solutions from diverse
trajectories occupy flatter minima and generalize better than any single trajectory.
HDET produces this diversity explicitly in parallel rather than sequentially.
The added communication cost over standard DDP is $O(1/T)$, negligible for $T \geq 100$.
In our production experiments (8 H100 GPUs, $T{=}1{,}000$), this extra AllReduce adds
approximately $0.1\%$ wall-clock time.

\subsection{Automatic Learning Rate Controller}
\label{sec:autolr}

After $S_{\mathrm{warm}}$ warmup steps, at each sync step $k$ the controller
computes per-rank losses $L_r^{(k)} = \frac{1}{T}\sum_t \ell(\theta_r^{(t)},
\mathcal{B}_r^{(t)})$, normalizes them against the mean $\bar{L}^{(k)}$ with
temperature $\sigma$, and softmax-weights each rank:
\begin{equation}
  s_r^{(k)} = \frac{\bar{L}^{(k)} - L_r^{(k)}}{\sigma}, \qquad
  w_r^{(k)} = \frac{\exp(s_r^{(k)})}{\sum_{r'}\exp(s_{r'}^{(k)})}.
  \label{eq:score}
\end{equation}
The deviation of the performance-weighted average LR from the unweighted mean is
the \emph{hypergradient} signal:
\begin{align}
  \tilde\eta^{(k)} &= \textstyle\sum_r w_r^{(k)}\eta_r^{(kT)}, \quad
  \bar\eta^{(k)} = \tfrac{1}{N}\textstyle\sum_r \eta_r^{(kT)}, \label{eq:wavg}\\
  \Delta^{(k)} &= \tilde\eta^{(k)} - \bar\eta^{(k)}. \label{eq:delta}
\end{align}
$\Delta > 0$ when higher-LR replicas outperform, signaling the base LR should rise.
A momentum velocity smooths the update:
\begin{equation}
  v^{(k+1)} = \beta\, v^{(k)} + (1-\beta)\,\Delta^{(k)}, \qquad
  \bar\eta^{(k+1)} = \bar\eta^{(k)}(1-\gamma) + v^{(k+1)}\lambda.
  \label{eq:velocity}
\end{equation}
where $\gamma$ is set to match OneCycleLR's decay floor over $K{=}(S{-}S_{\mathrm{warm}})/T$
remaining sync intervals:
\begin{equation}
  \gamma = 1 - \exp\!\left(-\frac{\log(f_{\mathrm{div}}/f_{\mathrm{final}})}{K}\right).
  \label{eq:decay}
\end{equation}

\paragraph{Random reassignment across hyperparameters.}
After updating $\bar\eta^{(k+1)}$, a fresh random permutation $\pi$ re-assigns
multipliers to ranks:
\begin{equation}
  \eta_r^{(k+1)} = \bar\eta^{(k+1)} \cdot \rho_{\pi(r)}.
  \label{eq:reassign}
\end{equation}
This is critical when exploring \emph{multiple} hyperparameters simultaneously
(e.g.\ per-group LRs).  Fixed rank-to-value assignments would create spurious
cross-parameter correlations that confound the hypergradient signal; drawing an
independent permutation $\pi^{(j)}$ per hyperparameter $j$ decorrelates them,
allowing the controller to track whichever hyperparameter most explains rank-loss
variance at each sync.  In this paper we demonstrate simultaneous exploration across
four optimizer parameter groups (\texttt{embedding}, \texttt{no\_decay},
\texttt{non\_embedding}, \texttt{transformer}), whose emergent differential
decay rates are analyzed in \S\ref{sec:discussion}.

\subsection{Warm Noisy Initialization}
\label{sec:warminit}

HDET can optionally be seeded from a pre-trained checkpoint rather than from random
initialization.  Before the first fan-out phase, independent Gaussian noise is added
to each parameter:
\begin{equation}
  \theta_r \;\leftarrow\; \bar\theta_0 + \epsilon_r, \qquad
  \epsilon_r \sim \mathcal{N}\!\left(0,\;\frac{\nu^2\,\|\bar\theta_0\|^2}{d}\right),
  \label{eq:warminit}
\end{equation}
where $\bar\theta_0$ is the pre-trained checkpoint, $d$ is the total number of
parameters, and $\nu$ controls the perturbation scale (we use $\nu{=}0.01$
throughout).  This initializes each replica in a distinct neighborhood of a
high-quality solution, providing a warm start that accelerates early convergence and
encourages replicas to explore different regions of the loss landscape near the
checkpoint.

Warm initialization \emph{alone} is insufficient to stabilize training at aggressive
learning rates: without periodic weight averaging, noisy weights compound with
high-LR updates and eventually diverge (the \textbf{Warm-Init} ablation in
\S\ref{sec:experiments} confirms this).  Its role is to supply better-quality
starting points for the ensemble, not to replace the stabilization provided by the
fan-out/converge cycle.  When HDET weight averaging is active, warm-init and the
auto-LR controller each contribute independently, as the ablation results demonstrate.

\begin{algorithm}[t]
\caption{Hyperparameter-Divergent Ensemble Training (HDET)}
\label{alg:hdet}
\begin{algorithmic}[1]
\Require Model $\theta$, optimizer, total steps $S$, world size $N$,
         OneCycleLR params ($\eta_{\max}$, $f_{\mathrm{div}}$, $f_{\mathrm{final}}$, $p$),
         spread ratio $\alpha$, sync interval $T$,
         warmup steps $S_{\mathrm{warm}}$, auto-LR params $(\beta, \sigma, \lambda)$
\State \textbf{Init:} $v \leftarrow \mathbf{0}$;\; $L_{\mathrm{acc}} \leftarrow 0$
\State Set $\rho_r$ via Eq.~\eqref{eq:rho};\;
       scale \texttt{base\_lrs} $\leftarrow$ \texttt{base\_lrs} $\times \rho_r$ on rank $r$
\For{$t = 1, \ldots, S$}
  \State Forward + backward; AllReduce gradients $\bar{g}^{(t)}$ (standard DDP)
  \State Compute $\eta_r^{(t)}$ from OneCycleLR formula with scaled \texttt{base\_lrs}
  \State Update: $\theta_r \leftarrow \theta_r - \eta_r^{(t)} \cdot \bar{g}^{(t)}$
  \State $L_{\mathrm{acc}} \leftarrow L_{\mathrm{acc}} + \ell(\theta_r, \mathcal{B}_r^{(t)})$
  \If{$t \bmod T = 0$}
    \State \textbf{Converge:} AllReduce $\theta_r \leftarrow \frac{1}{N}\sum_{r'}\theta_{r'}$
          \hfill (Eq.~\ref{eq:avg})
    \If{auto-LR enabled \textbf{and} $t \geq S_{\mathrm{warm}}$}
      \State $L_r \leftarrow L_{\mathrm{acc}}/T$;\; $L_{\mathrm{acc}} \leftarrow 0$
      \State AllGather $\{L_r\}_{r=0}^{N-1}$;\; compute $\{w_r\}$ via
             Eq.~\eqref{eq:score}
      \State AllGather $\{\eta_r^{(t)}\}_{r=0}^{N-1}$;\; compute
             $\tilde\eta$, $\bar\eta$, $\Delta$ via Eqs.~\eqref{eq:wavg}--\eqref{eq:delta}
      \State $v \leftarrow \beta v + (1-\beta)\Delta$
             \hfill (Eq.~\ref{eq:velocity})
      \State $\bar\eta \leftarrow \bar\eta(1-\gamma) + v \cdot \lambda$
             \hfill (Eq.~\ref{eq:velocity})
      \State Sample permutation $\pi$\ for random hyperparameter combinations; set
             $\eta_r \leftarrow \bar\eta \cdot \rho_{\pi(r)}$
             \hfill (Eq.~\ref{eq:reassign})
    \Else
      \State Update $\eta_r^{(t)}$ using standard OneCycleLR formula
    \EndIf
  \EndIf
\EndFor
\State \Return $\theta_r$ (all ranks identical at last sync step)
\end{algorithmic}
\end{algorithm}

\begin{remark}[Generalization to arbitrary hyperparameters]
\label{rem:general}
Algorithm~\ref{alg:hdet} is presented for learning rate $\eta$ but applies to any
scalar hyperparameter $h$ satisfying: (i)~varying $h$ across replicas does not
change model architecture (so AllReduce averaging remains well-defined), and
(ii)~$h$ influences training loss differentially (so loss comparisons carry signal
about $h$'s optimal value).
Candidates include dropout rate, label-smoothing coefficient, weight decay, and
attention temperature.
In each case $\Delta = \tilde{h} - \bar{h}$ is a zero-order hypergradient
estimating $-\partial\mathcal{L}/\partial h$ from inter-replica loss differences,
without back-propagating through the training loop.
Multiple hyperparameters can be co-explored simultaneously by assigning each replica
an independent spread vector, with a separate random permutation per hyperparameter
to prevent cross-parameter correlation.
\end{remark}

\section{Experiments}
\label{sec:experiments}

\subsection{Experimental Setup}

\paragraph{Task and data.}
We evaluate on a production-scale news-feed dataset from a major
social network, comprising one year of user--item interaction logs.
Models are trained jointly on three engagement tasks (\textbf{Contribution},
\textbf{Like}, \textbf{LongDwell}) for \textbf{one epoch}---standard in
production settings where the data stream is non-repeating and any change
in training loss directly reflects optimization quality.

\paragraph{Backbone.}
All experiments share the same fixed \textbf{AttnMVP} backbone
\citep{cheng2026} with identical architecture, optimizer (Adam), and
feature set.  Backbone details are orthogonal to HDET and are reported
in \citet{cheng2026}; our focus is solely on the training dynamics.

\paragraph{HDET configuration.}
All variants run on \textbf{8 NVIDIA H100 GPUs} ($N{=}8$); each single-epoch run
takes ${\approx}30$ hours, constraining ablation breadth.
Spread $\alpha{=}\nicefrac{1}{9}$, giving multipliers spanning $[0.0008,\,0.001]$
with mean $0.0009$ (matching the baseline).
$T{=}1{,}000$ steps; $S_{\mathrm{warm}}{=}100{,}000$ steps;
auto-LR: $\beta{=}0.9$, $\sigma{=}0.1$, $\lambda{=}0.5$ (fixed, not tuned).

\paragraph{Ablation configurations.}
We compare six configurations that isolate each HDET component:

\begin{enumerate}
  \item \textbf{Baseline-Low} ($\eta{=}0.0001$, no spread): conservative-LR reference.
  \item \textbf{Baseline-High} ($\eta{=}0.0009$, no spread): high-LR without ensemble stabilization.
  \item \textbf{Warm-Init} ($\eta{=}0.0009$, no spread): warm noisy initialization
        from a pre-trained checkpoint (\S\ref{sec:warminit}), no weight averaging.
        Tests whether initialization alone is sufficient for high-LR stability.
  \item \textbf{HDET w/o auto-LR} ($\eta{=}0.0009$, spread + warm init):
        fan-out/converge with LR spread and weight averaging; auto-LR disabled.
  \item \textbf{HDET w/o warm init} ($\eta{=}0.0009$, spread + auto-LR):
        full HDET from a cold start; ablates warm initialization.
  \item \textbf{HDET (full)} ($\eta{=}0.0009$, spread + warm init + auto-LR):
        complete method.
\end{enumerate}

We report average training loss across the three tasks (lower is better);
\textdagger~denotes training crash.

\subsection{Main Results}

\begin{table}[t]
\centering
\caption{%
  Training loss on production news-feed data.
  Lower is better.  \textdagger~indicates training crash (loss grew
  unboundedly).  All HDET variants use $\eta_{\max}{=}0.0009$, which is $9\times$
  higher than the conservative baseline.
}
\label{tab:main}
\begin{tabular}{llcc}
\toprule
\textbf{Model} & \textbf{Description} & \textbf{Avg LR} & \textbf{Train Loss} \\
\midrule
Baseline-Low      & Standard DDP, $\alpha{=}0$               & 0.0001 & 3.294 \\
Baseline-High     & Standard DDP, $\alpha{=}0$               & 0.0009 & 4.169\textdagger \\
Warm-Init         & Perturbed init, $\alpha{=}0$             & 0.0009 & 4.674\textdagger \\
\midrule
HDET w/o auto-LR  & LR spread + warm init, no auto-LR        & 0.0009 & 3.280 \\
HDET w/o warm init & LR spread + auto-LR, cold start         & 0.0009 & 3.281 \\
\textbf{HDET (full)} & LR spread + warm init + auto-LR       & 0.0009 & \textbf{3.277} \\
\bottomrule
\end{tabular}
\end{table}

\begin{figure}[t]
\centering
\includegraphics[width=\linewidth]{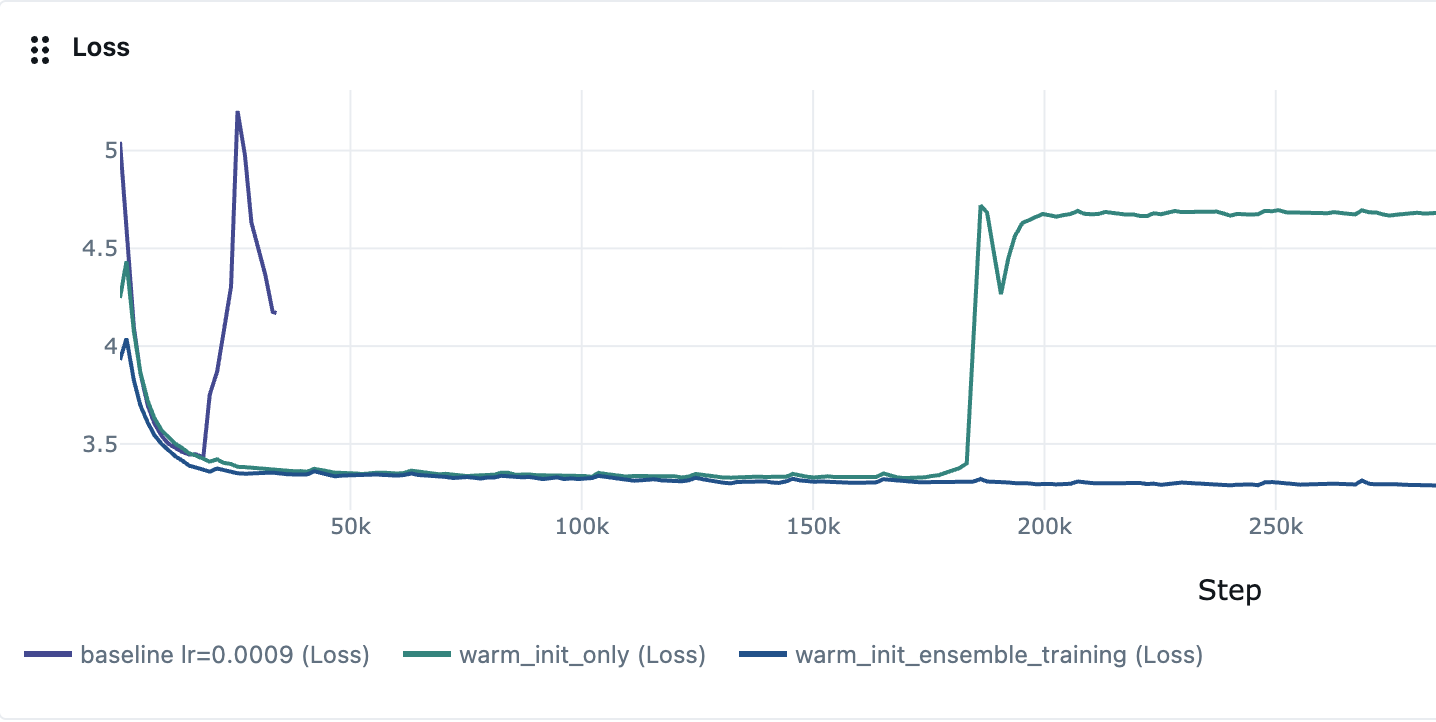}
\caption{%
  Training loss vs.\ steps at $\eta_{\max}{=}0.0009$.
  \textbf{Baseline} (gray) crashes at $\approx$18K steps, stabilizing at 4.169.
  \textbf{Warm-Init} (teal) survives to $\approx$185K steps then diverges (4.674):
  warm initialization extends stability but cannot replace weight averaging.
  \textbf{HDET} (dark blue) converges smoothly to 3.277---below Baseline-Low
  (3.294, not shown) trained at $9\times$ lower LR.
}
\label{fig:loss_curves}
\end{figure}

\begin{figure}[t]
\centering
\includegraphics[width=\linewidth]{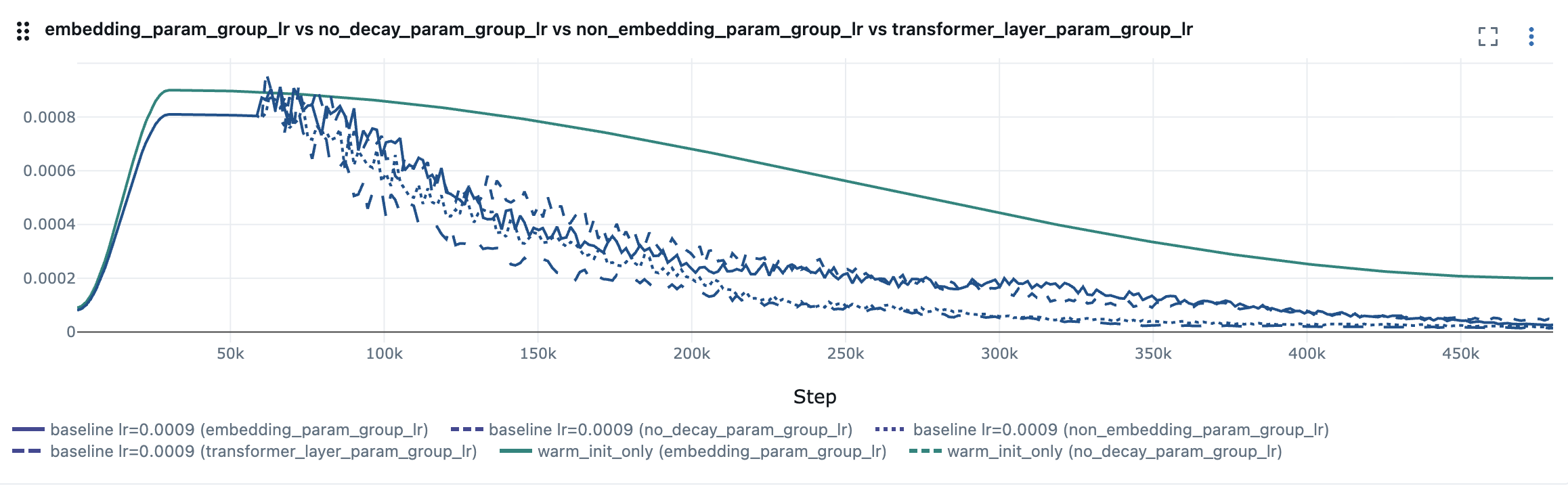}
\caption{%
  Per-group LR schedules for four parameter groups.
  Solid: baseline (OneCycleLR, $\eta_{\max}{=}0.0009$).
  Dashed: HDET GPU~0 (lowest-LR replica, $\rho_0{\cdot}0.0009{=}0.0008$);
  the 8-replica mean equals the $0.0009$ baseline.
  Periodic fluctuations mark each sync interval where the auto-LR controller
  updates the base schedule.
}
\label{fig:lr_schedules}
\end{figure}

\subsection{Analysis}

\paragraph{Stability (Figure~\ref{fig:loss_curves}).}
Both Baseline-High and Warm-Init diverge at $\eta{=}0.0009$; HDET trains stably
throughout.
Warm-Init diverges \emph{more} severely than Baseline-High (4.674 vs.\ 4.169),
confirming that weight averaging---not initialization---is the primary stabilizer.

\paragraph{Quality and ablation.}
HDET w/o auto-LR (3.280) already beats Baseline-Low (3.294) at $9\times$ higher LR,
showing that LR spread + weight averaging alone discovers better optima.
The two ablations---w/o auto-LR (3.280) and w/o warm init (3.281)---each contribute
independently and comparably (${\approx}0.013$--$0.014$ each), and HDET (full, 3.277)
combines both for the best result.
The auto-LR gain is modest here because the initial schedule is already near-optimal;
larger gains are expected when the hand-tuned baseline is farther from optimal.

\paragraph{Statistical reliability.}
We repeated training over \textbf{10+ independent models}; HDET consistently
outperformed Baseline-Low on offline AUC and NE across all replications.
The model is deployed in the production ranking system.

\paragraph{Per-group LR schedules (Figure~\ref{fig:lr_schedules}).}
Figure~\ref{fig:lr_schedules} shows GPU~0 (peak LR $0.0008 = \rho_0 \cdot 0.0009$,
the lowest-LR replica) for clarity; the 8-replica mean equals the $0.0009$ baseline.
The auto-LR controller produces periodic fluctuations at each sync and drives all
groups to a residual LR of ${\approx}0.0002$ rather than near-zero,
reflecting its discovery that a non-trivial final LR benefits this task.

\section{Discussion}
\label{sec:discussion}

\paragraph{Exploration-exploitation in hyperparameter space.}
HDET implements a principled exploration-exploitation trade-off directly in
hyperparameter space.
The \emph{fan-out} phase is the exploration step: replicas diverge under a structured
spread of learning rates, simultaneously probing distinct regions of the loss landscape
that no single schedule could visit concurrently.
The \emph{converge} phase is the exploitation step: AllReduce averaging collapses the
ensemble into a single model that harvests the best of all explored trajectories,
re-anchoring parameters near the consensus of the full replica population.
Because the LR spread always keeps some replicas conservative, this consensus is
guaranteed to lie in a stable region---which is why periodic averaging prevents
divergence where a single aggressive LR cannot.
Warm-Init without HDET illustrates the same mechanism in reverse: with no converge
phase to re-anchor parameters, high-LR updates on perturbed weights compound
unchecked until irrecoverable divergence at $\sim$185K steps.

\paragraph{Emergent differential LR decay.}
The auto-LR controller discovers distinct per-group decay rates without manual
specification (Figure~\ref{fig:lr_schedules}):
\texttt{transformer} decays fastest (dense high-variance gradients converge quickly),
\texttt{non\_decay} parameters slowest (flat landscapes absorb gradient signal
throughout), with embeddings and non-embedding layers in between.
This ordering mirrors discriminative fine-tuning \citep{howard2018universal} but
emerges \emph{automatically} from cross-replica loss signals.

\paragraph{Limitations.}
HDET's auto-LR adapts per-group schedules from model performance across replicas,
complementing Adam's \citep{kingma2014adam} per-parameter gradient-moment adaptation; combining them is a
natural next step.
The three additional hyperparameters ($\alpha$, $T$, $\sigma$) were fixed without
tuning and consistently outperformed the baseline, though their interaction and
architecture-dependence warrant further study.

\section{Conclusion and Future Work}
\label{sec:conclusion}

HDET converts the wasted uniformity of DDP replicas into a structured LR
exploration ensemble via a fan-out/converge cycle ($O(1/T)$ extra communication),
with an auto-LR controller that adapts the base schedule online from cross-replica
loss signals and warm noisy initialization that seeds training near a high-quality
checkpoint.
On a production recommendation model, HDET trains stably with much higer LR than the LR that
causes divergence in standard DDP and achieves a lower final loss (3.277 vs.\
3.294), while its auto-LR controller autonomously discovers the empirically correct
per-group decay ordering (transformer fastest, non-decay slowest).

\paragraph{Future directions.}
\textit{(i) General hyperparameter exploration.}
As formalized in Remark ~\ref{rem:general}, any scalar hyperparameter that does not
change model architecture (dropout, weight decay, temperature) can be substituted
for LR, turning the cluster into a zero-cost hyperparameter search engine using
inter-replica losses as hypergradients.
\textit{(ii) Group-aware initialization.}
The emergent decay ordering provides an empirical prior for initializing per-group
spread ratios, potentially accelerating group-wise schedule discovery.
\textit{(iii) LLM pretraining.}
Extending HDET to multi-epoch pretraining over hundreds of billions of tokens
offers more sync intervals for the auto-LR to adapt, with warm-init generalized to
early-stage cold starts.
\textit{(iv) Convergence theory.}
A formal analysis characterizing the interplay of LR spread, sync interval, and
averaging on convergence and implicit regularization remains an open question.

\begin{ack}
The authors thank colleagues on the production ranking team for helpful
discussions and infrastructure support.  We also thank the anonymous reviewers
for their constructive feedback.
\end{ack}

\section*{References}

\bibliographystyle{abbrvnat}
\bibliography{base}


\appendix


\newpage

\end{document}